\documentclass{article}

\usepackage[final,dblblindworkshop]{neurips_2026}
\workshoptitle{2nd Workshop on Agentic AI Benchmarks and Applications for Enterprise Tasks (NeurIPS 2026). Correspondence emails should be sent to: jwillette@nvidia.com}
\usepackage[utf8]{inputenc}
\usepackage[T1]{fontenc}
\usepackage{xcolor}
\usepackage{hyperref}
\usepackage{url}
\usepackage{booktabs}
\usepackage{graphicx}
\usepackage{subcaption}
\usepackage{microtype}

\definecolor{LegendNemotronDark}{HTML}{3D6F00}
\definecolor{LegendQwenDark}{HTML}{3F3AA8}
\definecolor{LegendDeepSeek}{HTML}{4D6BFE}
\colorlet{PaperLinkColor}{LegendQwenDark!88!black}
\colorlet{PaperCiteColor}{LegendDeepSeek!78!black}
\colorlet{PaperUrlColor}{LegendNemotronDark!92!black}
\hypersetup{
  colorlinks=true,
  linkcolor=PaperLinkColor,
  citecolor=PaperCiteColor,
  urlcolor=PaperUrlColor,
  pdfborder={0 0 0},
  bookmarksnumbered=true
}

\newcommand{\NemotronNanoExactSixtyFiveK}{0.0}
\newcommand{\NemotronNanoExactCountSixtyFiveK}{0}
\newcommand{\NemotronSuperTwoK}{0.711}
\newcommand{\NemotronSuperSixtyFiveK}{0.138}
\newcommand{\NemotronSuperExactSixtyFiveK}{0.0}
\newcommand{\NemotronSuperExactCountSixtyFiveK}{0}

\newcommand{\QwenThirtyFiveExactSixtyFiveK}{0.8}
\newcommand{\QwenThirtyFiveExactCountSixtyFiveK}{2}
\newcommand{\QwenOneTwentyTwoTwoK}{0.883}
\newcommand{\QwenOneTwentyTwoSixtyFiveK}{0.441}
\newcommand{\QwenOneTwentyTwoExactSixtyFiveK}{4.6}
\newcommand{\QwenOneTwentyTwoExactCountSixtyFiveK}{11}
\newcommand{\DeepSeekFlashTwoK}{0.909}
\newcommand{\DeepSeekFlashSixtyFiveK}{0.554}
\newcommand{\DeepSeekFlashExactSixtyFiveK}{17.1}
\newcommand{\DeepSeekFlashExactCountSixtyFiveK}{41}

\newcommand{\KimiLinearExactSixtyFiveK}{0.0}
\newcommand{\KimiLinearExactCountSixtyFiveK}{0}

\newcommand{\FalconHOneExactSixtyFiveK}{0.0}
\newcommand{\FalconHOneExactCountSixtyFiveK}{0}
\newcommand{\MeanInputTokensSixtyFiveK}{65{,}374}
\newcommand{\MeanRequiredOutputTokensSixtyFiveK}{61{,}240}

\newcommand{\DeepSeekArithmeticSixtyFiveK}{0.961}
\newcommand{\DeepSeekTableSixtyFiveK}{0.262}
\newcommand{\QwenOneTwentyTwoArithmeticSixtyFiveK}{0.952}
\newcommand{\QwenOneTwentyTwoUUIDSixtyFiveK}{0.233}
\newcommand{\QwenOneTwentyTwoTableSixtyFiveK}{0.170}

\title{Staying on Task: Testing the Foundations of Long-Horizon Agent Reliability}

\author{%
  Jeffrey Willette \quad Krishna C. Puvvada \quad Boris Ginsburg \\
  NVIDIA\\
}

\begin{document}

\maketitle

\begin{abstract}
Long-horizon agentic workflows require models to sustain repeated state-dependent actions all while the context grows, sub-task complexity changes, and new data arrives. Each situation represents an independent axis along which an agent may fail. An agent reconciling a long ledger, for example, must repeatedly read its state, update the correct record, and preserve alignment across thousands of outputs. A model may accept the entire ledger yet lose its place or stop applying the operation consistently as generation proceeds. We introduce Long-Transduction, a controlled diagnostic that tests a model's ability to stay on task during long generation while continuously reading, mutating, and outputting input-context dependent operations such as arithmetic, sorting, variable lookups, and table transformations. Long-Transduction evaluation independently varies local task complexity, input data formatting, and context length isolate failures along each axis. We evaluate seven open-weight models, finding a 62.8\% decrease when scaling context length from 4-128K, a 36.5\% decrease when varying input format, and a 39.9\% decrease by increasing local task complexity. Together, these failures represent critical liabilities in long-horizon agentic workflows.
\end{abstract}


\section{Long-horizon Execution Needs Its Own Diagnostic}
Consider an accounts-payable agent reconciling a 20,000-line invoice export against purchase orders. For each line it must retrieve the right vendor terms, compute an adjustment, and update the matching ledger entry. If the final ledger is wrong, a task-level score cannot tell whether the agent retrieved the wrong terms, miscomputed the adjustment, dropped one invoice, or shifted every later decision onto the wrong record. Benchmarks spanning real-world questions, web interactions, code repair, and tool-mediated goals measure whether the overall task succeeded~\citep{mialon2024gaia,zhou2024webarena,drouin2024workarena,jimenez2024swebench,yao2024taubench}; they do not, however, isolate this sustained-execution failure in a way that can be measured.

Long-context benchmarks test complementary retrieval and understanding abilities~\citep{bai2024longbench,hsieh2024ruler,yen2025helmet}. RULER, for example, extends needle-in-a-haystack retrieval with multiple keys, tracking, and aggregation, but still produces answers short relative to input context. In a long horizon retrieval and transformation, the output itself is a long stateful trajectory that requires careful bookkeeping and attention to detail. Long-Transduction implements this paradigm by holding the atomic operations simple which provides a way to score every required item in the output. A model that cannot stay aligned while repeatedly executing these elementary operations cannot be expected to stay on task through more complex actions in a long-horizon agentic workflow such as Browsecomp~\citep{browsecomp} . We contribute (i) a controlled design that independently varies length, local difficulty, and input format; (ii) exact per-record and position-resolved scoring that reveals what fails and when during generation; and (iii) a balanced seven-model evaluation, showing that supported context length does not certify reliable completion.

\begin{figure}[t]
  \centering
  \makebox[\textwidth][c]{\includegraphics[width=1.03\textwidth]{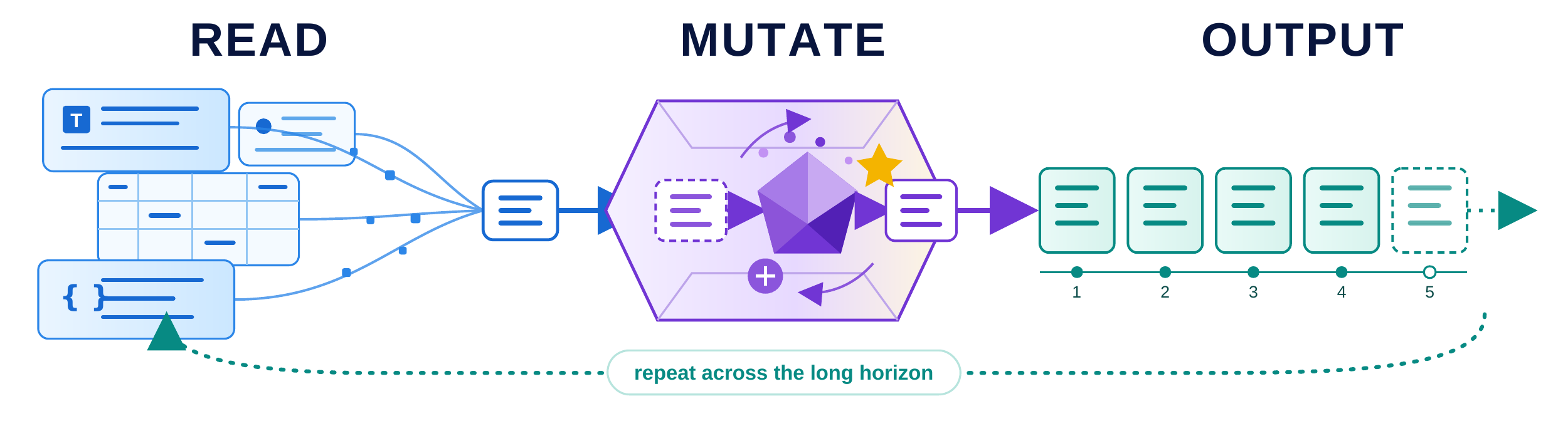}}
  \caption{The universal read, mutate, output loop underlying both transduction tasks and agentic workflows. At each step, a system reads relevant state from its context, applies a meaningful mutation, and appends the result to its output stream. Long-transduction tests whether models can repeat this loop reliably across a growing long-horizon workflow.}
  \label{fig:concept}
\vspace{-1em}
\end{figure}

\section{Long-transduction}
Figure~\ref{fig:concept} shows the read, mutate, output paradigm that Long-transduction isolates. We instantiate this loop with four controlled task families, three input formats, and four difficulty settings. The \emph{Arithmetic} task evaluates addition/subtraction expressions line-by-line and varies difficulty by including 2, 4, 8, or 16 operands per record. The \emph{UUID sorting} task sorts lists of identifiers line-by-line and varies difficulty by including 2, 4, 8, or 16 items per record. The \emph{Variable Lookup} task resolves key pairs against dictionaries line-by-line and varies difficulty by scaling total dicitionary size to 8, 32, 128, or 256 total entries. The \emph{Table Transformation} task applies a transformation to a CSV table and varies difficulty by increasing the complexity of the transformation. For examples of inputs from each task family as well as a more detailed explanation, please see~Appendix~\ref{app:examples}.

The first three families each have three matched input formats that test position tracking ability. First, ordered records with numeric IDs give every input line a stable lookup key as an integer index. Second, shuffling the input record ID's and requiring monotonically increasing ID's in the output adds a \textit{global} reordering requirement while preserving the \textit{local task} of a transformation operation on each row. Third, removing IDs provides the least structure: the model must dynamically track the context position based solely on the current output position and surrounding content. An omission could shift all future positions, leading to failure. Together, these settings evaluate local/global transformation ability, retrieval, and position tracking; all fundamental tasks in long horizon workflows.

Every one of the 4 tasks is evaluated with three different input formats, four difficulty settings, six context budgets, and five document samples. Thus, $12\times4\times6\times5=1{,}440$ documents per model, including 240 at each context length horizon. A single document can contain hundreds to thousands of individually scored items. Unless otherwise specified, we keep the document, not each item, as the independent scored unit. Because required output grows approximately one-for-one with input, we report nominal input-plus-output horizons of 4K, 8K, 16K, 32K, 64K, and 128K tokens. For example, a 4K transduction document is approximately 2K input tokens and 2K output tokens. At the largest tier, the data average \MeanInputTokensSixtyFiveK{} input and \MeanRequiredOutputTokensSixtyFiveK{} required output tokens.

Every reported item accuracy is the fraction of required output records within a document that are exactly correct. We average records within each document and then give every task-difficulty cell equal weight, so settings containing more records do not dominate.

\section{Open-weight Model Evaluation}
We test eight open-weight checkpoints: Nemotron 3  30B and Super 120B~\citep{nemotron3}, Qwen3.5 35B-A3B and 122B-A10B~\citep{qwen3}, DeepSeek V4 Flash~\citep{deepseekv4}, Kimi Linear 48B-A3B~\citep{kimilinear}, Falcon-H1 34B~\citep{falconh1}, and Olmo 3.1 32B~\citep{olmo3}. Seven span the full grid; Olmo stops at 32K nominal tokens (16K input) because of its context limit. Closed-weight models frequently declined the task, terminated early, or requested clarification in pilot runs, so we omit them from quantitative comparisons. All models receive identical instructions with greedy sampling. Prompts explicitly say ``Do not think.'' We disable thinking/reasoning when an `off' or `none' setting exists and otherwise use the lowest available thinking budget (Deepseek only); all stored rollouts report zero reasoning tokens. Model-specific tokenizers produce somewhat different observed input lengths (Appendix~\ref{sec:reproducibility}).

\begin{figure}[t]
  \centering
  \includegraphics[width=0.66\textwidth]{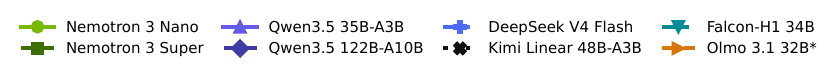}\\[-2pt]
  \begin{subfigure}[t]{0.425\textwidth}
    \centering
    \includegraphics[width=\linewidth]{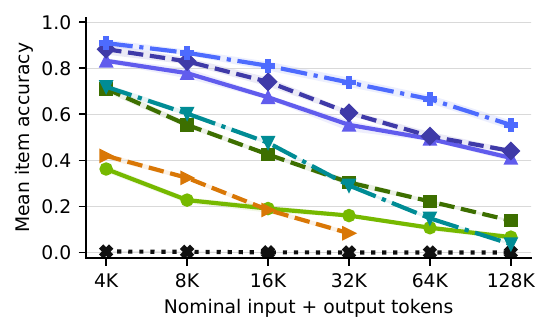}
    \caption{Reliability by context horizon.}
    \label{fig:main-horizon}
  \end{subfigure}\hfill
  \begin{subfigure}[t]{0.425\textwidth}
    \centering
    \includegraphics[width=\linewidth]{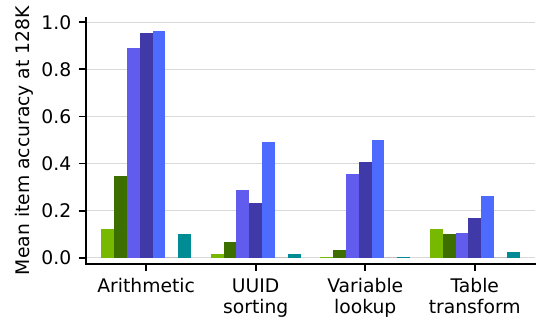}
    \caption{Individual task performance at 128K.}
    \label{fig:main-families}
  \end{subfigure}
  \caption{Fig.~\ref{fig:main-horizon} shows how accuracy degrades as context length increases. The result is averaged over all tasks, input formats, and difficulties. Fig.~\ref{fig:main-families} shows individual task performance at 128K averaged over input formats and difficulties. Olmo* ends at 32K because of its context limit.}
  \label{fig:main}
\vspace{-1em}
\end{figure}

\section{Results}
\paragraph{Horizon exposes large reliability losses.}
Figure~\ref{fig:main}\subref{fig:main-horizon} shows monotonic degradation for every non-degenerate model. DeepSeek leads throughout, but overall accuracy falls from \DeepSeekFlashTwoK{} at 4K nominal tokens to \DeepSeekFlashSixtyFiveK{} at 128K. Qwen-122B falls from \QwenOneTwentyTwoTwoK{} to \QwenOneTwentyTwoSixtyFiveK{}, only 0.030 above Qwen-35B at 128K. Nemotron Super falls from \NemotronSuperTwoK{} to \NemotronSuperSixtyFiveK{}. Falcon falls from 0.719 to 0.036 and Olmo from 0.419 to 0.083 (at 32K). Supported context and model scale therefore do not certify reliable exhaustive execution. \textbf{Averaged over all models, there is a 62.8\% relative decrease in performance when scaling context from 4K to 128K.}

\paragraph{Individual task reliability.}
Task-family results diverge sharply (Fig.~\ref{fig:main}\subref{fig:main-families}): at 128K nominal tokens, DeepSeek reaches \DeepSeekArithmeticSixtyFiveK{} on arithmetic but only \DeepSeekTableSixtyFiveK{} on table transformation; Qwen-122B combines \QwenOneTwentyTwoArithmeticSixtyFiveK{} arithmetic with \QwenOneTwentyTwoUUIDSixtyFiveK{} UUID sorting and \QwenOneTwentyTwoTableSixtyFiveK{} table accuracy. For exhaustive workflows, an incorrect or missing record forces reconciliation or retry, so exact completion estimates the fraction requiring no repair. At 128K this is only \DeepSeekFlashExactCountSixtyFiveK{}/240 for DeepSeek (\DeepSeekFlashExactSixtyFiveK{}\%), \QwenOneTwentyTwoExactCountSixtyFiveK{}/240 for Qwen-122B (\QwenOneTwentyTwoExactSixtyFiveK{}\%), and \QwenThirtyFiveExactCountSixtyFiveK{}/240 for Qwen-35B (\QwenThirtyFiveExactSixtyFiveK{}\%); Appendix Table~\ref{tab:exact-completion} reports all complete-grid models.

\begin{figure}[t]
  \centering
  \includegraphics[width=0.66\textwidth]{figures/legend_models.pdf}\\[-2pt]
  \begin{subfigure}[t]{0.32\textwidth}
    \centering
    \includegraphics[width=\linewidth]{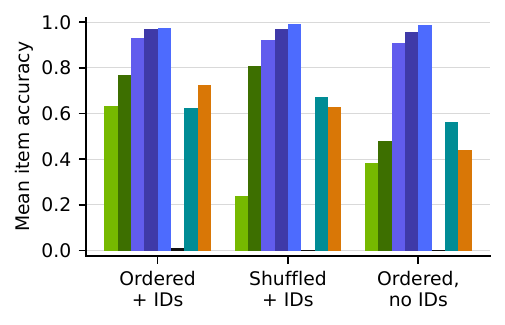}
    \caption{Arithmetic}
    \label{fig:protocol-arithmetic}
  \end{subfigure}\hfill
  \begin{subfigure}[t]{0.32\textwidth}
    \centering
    \includegraphics[width=\linewidth]{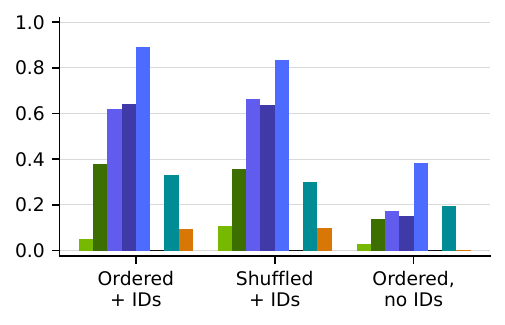}
    \caption{UUID sorting}
    \label{fig:protocol-uuid}
  \end{subfigure}\hfill
  \begin{subfigure}[t]{0.32\textwidth}
    \centering
    \includegraphics[width=\linewidth]{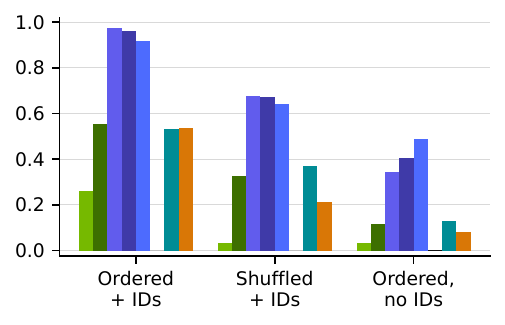}
    \caption{Variable lookup}
    \label{fig:protocol-lookup}
  \end{subfigure}
  \caption{Format sensitivity at 128K nominal tokens, averaged over four difficulties. Surprisingly, ordered transformations without row ID's scores worse than shuffling the numbered input rows. Removing IDs removes the stable row lookup key. Olmo* only includes up to 32K. For the corresponding table transform plot, see Fig.~\ref{fig:protocol-table}.}
  \label{fig:protocol}
\vspace{-1em}
\end{figure}

\paragraph{Format reveals conditional failures.}
Fig.~\ref{fig:protocol-arithmetic} shows a level of robustness to changes in the input format. Averaged over all models, there is a 16.2\% decrease when moving from `Ordered + IDs' to `Ordered, No IDs' However, for Figs.~\ref{fig:protocol}\subref{fig:protocol-uuid}-\subref{fig:protocol-lookup}, we see that UUID sorting and variable lookup are much more sensitive to input format, with the unstructured `no ID' variant realizing a 64.3\% and 66.4\% decrease respectively. \textbf{This indicates that the ability to track a content position in the context is not invariant to the current task which is being performed.}


\begin{figure}[t]
  \centering
  \includegraphics[width=0.66\textwidth]{figures/legend_models.pdf}\\[-2pt]
  \begin{subfigure}[t]{0.32\textwidth}
    \centering
    \includegraphics[width=\linewidth]{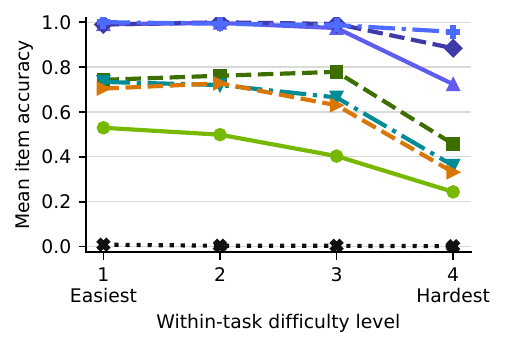}
    \caption{Arithmetic}
    \label{fig:overall-difficulty-column1}
  \end{subfigure}\hfill
  \begin{subfigure}[t]{0.32\textwidth}
    \centering
    \includegraphics[width=\linewidth]{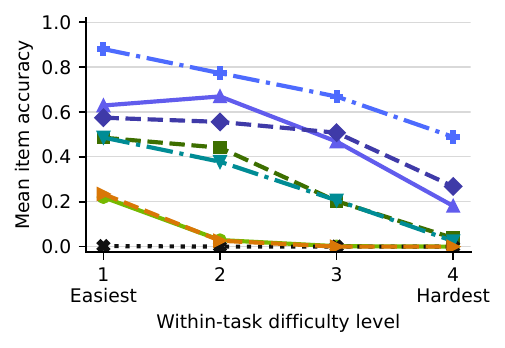}
    \caption{UUID sort}
    \label{fig:overall-difficulty-column2}
  \end{subfigure}\hfill
  \begin{subfigure}[t]{0.32\textwidth}
    \centering
    \includegraphics[width=\linewidth]{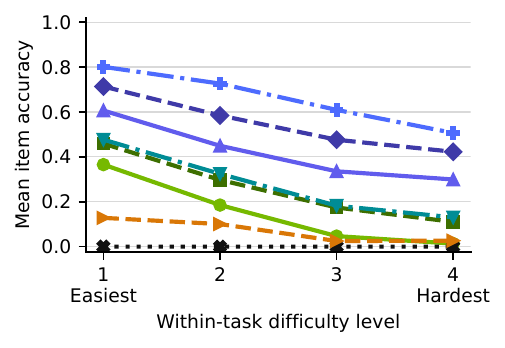}
    \caption{Table Transform}
    \label{fig:overall-difficulty-column3}
  \end{subfigure}
  \caption{Overall Accuracy at 128K decreases with an increasing local task complexity. Metrics are averaged over input format. Increasing local task complexity degrades overall performance. Olmo* data only up until 32K. For the corresponding variable lookup plot, see Fig.~\ref{fig:variable-lookup-difficulty}}
  \label{fig:overall-difficulty}
\vspace{-1em}
\end{figure}

\paragraph{Local difficulty.}
Fig.~\ref{fig:overall-difficulty}\subref{fig:overall-difficulty-column1}-\subref{fig:overall-difficulty-column3} show the effect of adding more local task complexity by increasing summands, sorting items, or permuted rows/columns. Every non-degenerate model declines as local work increases from the easiest to hardest. Appendix Fig.~\ref{fig:difficulty} expands these marginals into all 12 task-specific curves at 128K and includes the arithmetic task.


\begin{figure}[!ht]
  \vspace{-1em}
  \centering
  \includegraphics[width=0.66\textwidth]{figures/legend_models.pdf}\\[-2pt]
  \begin{subfigure}[t]{0.485\textwidth}
    \centering
    \includegraphics[width=\linewidth]{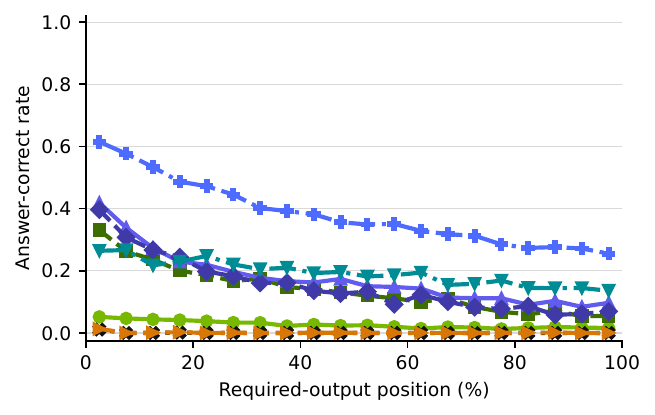}
    \caption{Answer Correct}
    \label{fig:answer-correct}
  \end{subfigure}\hfill
  \begin{subfigure}[t]{0.485\textwidth}
    \centering
    \includegraphics[width=\linewidth]{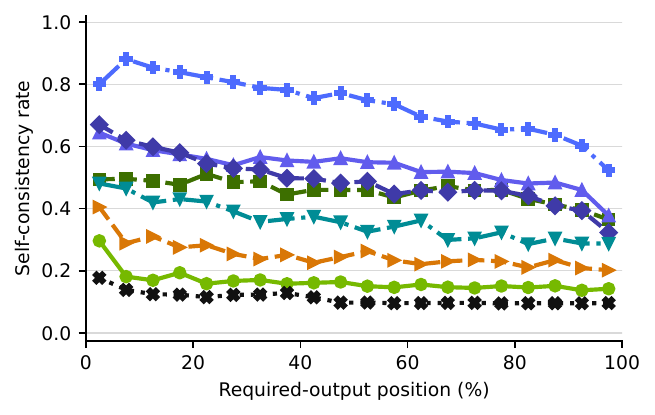}
    \caption{Self-consistency}
    \label{fig:self-consistent}
  \end{subfigure}
  \caption{Calculating individual scored items throughout generation for the `Ordered, no IDs' format on the UUID sorting task averaged over all context lengths and difficulties. Models tend to stay self-consistent longer than they are able to copy information from context.}
  \label{fig:position}
\vspace{-1em}
\end{figure}

\paragraph{Models understand the task, but get lost.}
Due to the synthetic structure of the task, we are able to score individual items in the output stream as a function of output position. Then, we can ask an interesting quesion: Even if the model didn't output the correct answer, did it at least stay self-consistent? That is, for UUID sorting, did it output a properly sorted list of `something?' Figs.\ref{fig:answer-correct} and \ref{fig:self-consistent} show that self-consistency is always significantly higher than per-position accuracy, indicating the model understands the task, but fails to read the correct problem from the context. 

\section{Conclusion}
\label{sec:conclusion}
These results have important implications for long-horizon agents. Long-transduction shows that nominal capacity can hide failures caused by input formats, local difficulty, and context length. Currently, long workflows should be sure to itemize inputs with stable IDs, checkpoint chunks of outputs with IDs, and split tasks into small and less complex units of work to avoid presenting difficult tasks in conjunction with long context and confusing input formats. Furthermore, future models should focus more effort on mitigating these failure modes at train-time so that agents may handle complex, less-structured, long-horizon tasks.

\label{main:end}
\clearpage
\bibliographystyle{plainnat}
\bibliography{references}

\clearpage
\appendix
\section{Benchmark details and examples}
\label{app:examples}

\paragraph{Arithmetic.}
The ordered numbered variant isntructs the model to transform the following input:
\begin{center}\small\begin{tabular}{l}
\texttt{[1]5+6}\\
\texttt{[2]2+4-1}\\
\texttt{[3]4+8-3+2}
\end{tabular}\end{center}
into the following output:
\begin{center}\small\begin{tabular}{l}
\texttt{[1]5+6=11}\\
\texttt{[2]2+4-1=5}\\
\texttt{[3]4+8-3+2=11}.
\end{tabular}\end{center}
The shuffled variant presents the same input records with shuffled rows, and requires an ascending (sorted) output index; the unnumbered variant removes the row index identifiers (such as \texttt{[1], [2], [3]} and requires the model to keep track of the global position without an index to use as a lookup key. The four difficulty levels are created by making each summand consist of 2,4,8, or 16 terms.

\paragraph{UUID sorting.}
Each record contains short lists of hexadecimal identifiers:
\begin{center}\small\begin{tabular}{l}
\texttt{[1]c0a8e1d2,a1b2c3d4,b1c2d3e4}\\
\texttt{[2]f0e1d2c3,01234567}
\end{tabular}\end{center}
and the model must output the same list in sorted order:
\begin{center}\small\begin{tabular}{l}
\texttt{[1]a1b2c3d4,b1c2d3e4,c0a8e1d2}\\
\texttt{[2]01234567,f0e1d2c3}.
\end{tabular}\end{center}
The verifier marks the record correct only when the emitted identifiers exactly match the expected sorted sequence. Similar to arithmetic, this task includes Ordered, Shuffled, and No ID variants. The four levels of difficulty are created by including 2,4,8, or 16 items in each hexadecimal list. 

\paragraph{Variable lookup.}
A variable definition pool such as
\begin{center}\small\begin{tabular}{ll}
\texttt{a3f=quick} & \texttt{b91=fox}\\
\texttt{03c=quiet} & \texttt{e18=harbor}
\end{tabular}\end{center}

is followed by expressions such as:

\begin{center}\small\begin{tabular}{l}
\texttt{[1]a3f+b91} \\
\texttt{[2]o3c+e18} \\
\end{tabular}\end{center}

The expected output transformation of each record is:

\begin{center}\small\begin{tabular}{l}
\texttt{[1]quick fox} \\
\texttt{[2]quiet harbor} \\
\end{tabular}\end{center}

Like the arithmetic and UUID variants, variable lookup also includes ordered, shuffled, and no ID variants. The four difficulty settings are achieved by changing the definition-pool size to one of 8, 32, 128, or 256 independently of the number of output records.

\paragraph{CSV Table Transformation.}
The table transformation task follows a slightly different pattern than the previous variants. For an input record corresponding to a CSV table
\begin{center}\small\begin{tabular}{l}
\texttt{,[C0],[C1],[C2]}\\
\texttt{[R0],a,b,c}\\
\texttt{[R1],d,e,f}\\
\texttt{[R2],g,h,i}
\end{tabular}\end{center}

CSV tables do not have a clear input format that can be indexed by row, as the previous tasks do. Therefore, we construct the three analogous format variants by supplying instructions to perform an operation on either the rows and columns or the individual cells of the table. The operations are as follows:

\begin{enumerate}
    \item Row/Column permutation with homogeneous cells: each cell contains homogeneous length four-digit integers and the rows and columns must be permuted according to a given reordering.
    \item Row/Column permutation with heterogeneous cells: each cell contains variable-length UUID fragments ranging in length from 1-36 hexadecimal digits and the rows and columns must be permuted according to a given reordering.. 
    \item KV resolution: each cell contains a variable expression that needs to be de referenced. See the description below.
\end{enumerate}

For table transformation tasks requiring a row/column permutation, the model receives a new row and column order such as \texttt{[R2],[R0],[R1]} and \texttt{[C1],[C0],[C2]}, the output is then expected to be:

\begin{center}\small\begin{tabular}{l}
\texttt{,[C1],[C0],[C2]}\\
\texttt{[R2],h,g,i}\\
\texttt{[R0],b,a,c}\\
\texttt{[R1],e,d,f}.
\end{tabular}\end{center}

CSV tasks with permutation increase difficulty by requiring one of 20\%, 40\%, 80\%, or 100\% of rows and columns permuted.

\paragraph{CSV KV resolution.}
Given adjective and noun tables (e.g., \texttt{a0=quick}, \texttt{n0=fox}) and cells containing \texttt{a0+n0}, the model must preserve the CSV structure while replacing every cell with the resolved phrase. Difficulty and context length control how many distinct adjective and noun keys are defined by variables as displayed in the following table:

\begin{center}
\captionof{table}{Number of active adjective or noun variables in each dictionary. A value of `2' means that there are 2 adjectives and 2 nouns.}
\label{tab:csv-kv-vocab}
\begin{tabular}{c rrrr}
    \toprule
     Tokens / Difficulty & 1 & 2 & 3 & 4 \\
    \midrule
    2,048  & 2  & 5  & 10 & 19  \\
    4,096  & 4  & 7  & 14 & 28  \\
    8,192  & 5  & 10 & 21 & 42  \\
    16,384 & 8  & 15 & 30 & 61  \\
    32,768 & 11 & 22 & 44 & 87  \\
    65,536 & 16 & 31 & 62 & 125 \\     
    \bottomrule
\end{tabular} 
\end{center}

\section{Scoring}
Each required output record (output row or csv cell) is marked correct only when it exactly matches the expected answer. We first average record correctness within a generated document, then average equally over other dimensions such as tasks, difficulty, format, or context length. Exact document completion is an indicator equal to one only when every required record in that document is correct; the reported rate in Table~\ref{tab:exact-completion} averages this indicator over the 240 documents at the 128K horizon.

\begin{figure}[p]
  \centering
  \captionsetup[subfigure]{font=scriptsize,skip=1pt}
  \includegraphics[width=0.82\textwidth]{figures/legend_models.pdf}\\[-3pt]
  \begin{subfigure}[t]{0.32\textwidth}
    \centering\includegraphics[width=\linewidth]{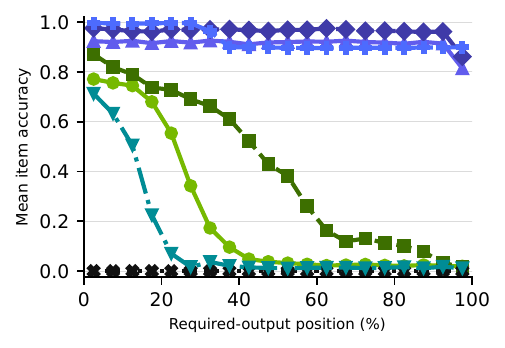}
    \caption{Arithmetic: ordered + IDs}\label{fig:position-all-arithmetic-ordered}
  \end{subfigure}\hfill
  \begin{subfigure}[t]{0.32\textwidth}
    \centering\includegraphics[width=\linewidth]{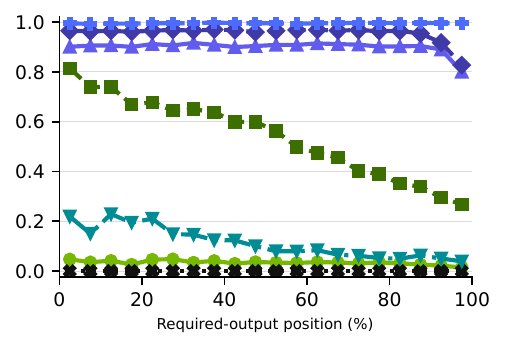}
    \caption{Arithmetic: shuffled + IDs}\label{fig:position-all-arithmetic-shuffled}
  \end{subfigure}\hfill
  \begin{subfigure}[t]{0.32\textwidth}
    \centering\includegraphics[width=\linewidth]{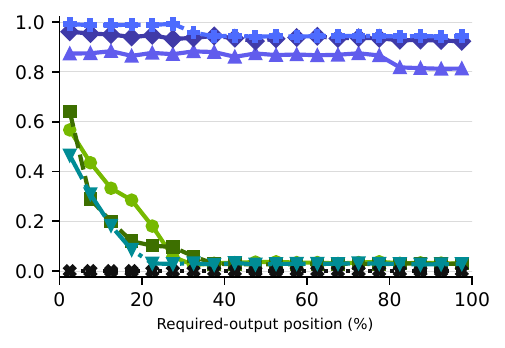}
    \caption{Arithmetic: no IDs}\label{fig:position-all-arithmetic-noids}
  \end{subfigure}

  \begin{subfigure}[t]{0.32\textwidth}
    \centering\includegraphics[width=\linewidth]{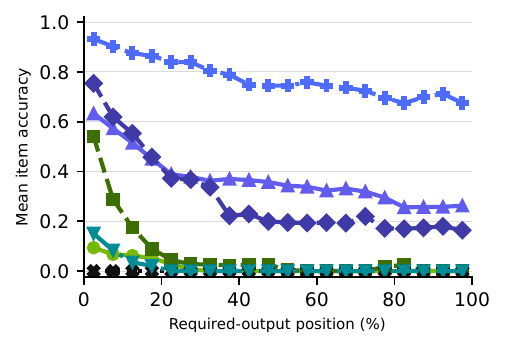}
    \caption{UUID sorting: ordered + IDs}\label{fig:position-all-uuid-ordered}
  \end{subfigure}\hfill
  \begin{subfigure}[t]{0.32\textwidth}
    \centering\includegraphics[width=\linewidth]{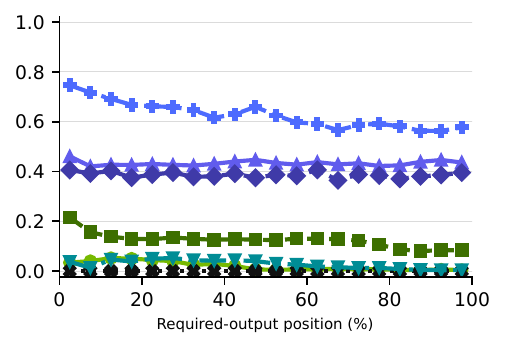}
    \caption{UUID sorting: shuffled + IDs}\label{fig:position-all-uuid-shuffled}
  \end{subfigure}\hfill
  \begin{subfigure}[t]{0.32\textwidth}
    \centering\includegraphics[width=\linewidth]{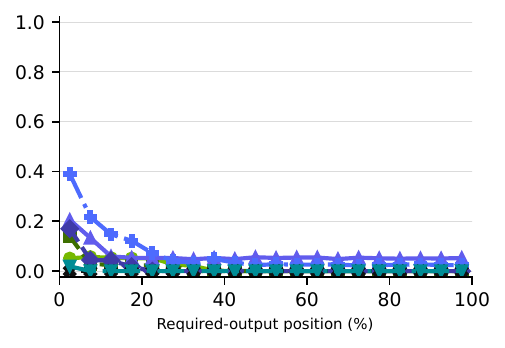}
    \caption{UUID sorting: no IDs}\label{fig:position-all-uuid-noids}
  \end{subfigure}

  \begin{subfigure}[t]{0.32\textwidth}
    \centering\includegraphics[width=\linewidth]{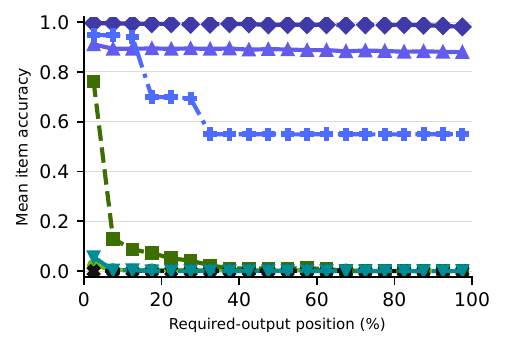}
    \caption{Lookup: ordered + IDs}\label{fig:position-all-lookup-ordered}
  \end{subfigure}\hfill
  \begin{subfigure}[t]{0.32\textwidth}
    \centering\includegraphics[width=\linewidth]{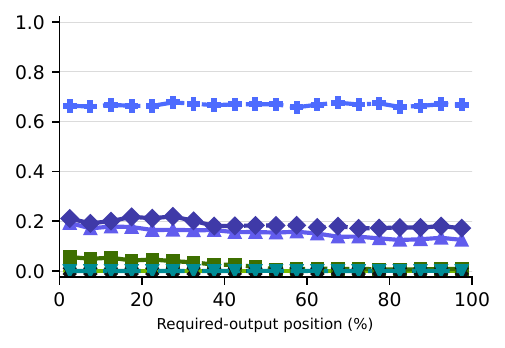}
    \caption{Lookup: shuffled + IDs}\label{fig:position-all-lookup-shuffled}
  \end{subfigure}\hfill
  \begin{subfigure}[t]{0.32\textwidth}
    \centering\includegraphics[width=\linewidth]{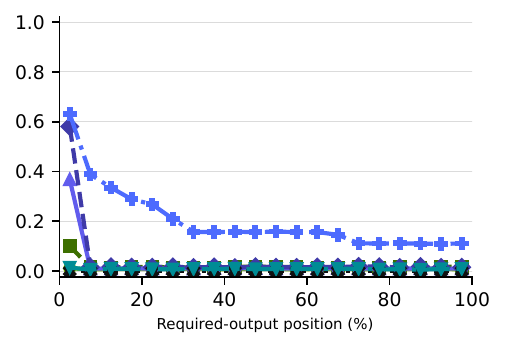}
    \caption{Lookup: no IDs}\label{fig:position-all-lookup-noids}
  \end{subfigure}

  \begin{subfigure}[t]{0.32\textwidth}
    \centering\includegraphics[width=\linewidth]{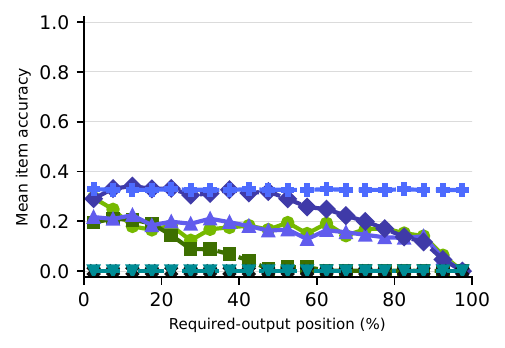}
    \caption{Homogeneous table permutation}\label{fig:position-all-table-homogeneous}
  \end{subfigure}\hfill
  \begin{subfigure}[t]{0.32\textwidth}
    \centering\includegraphics[width=\linewidth]{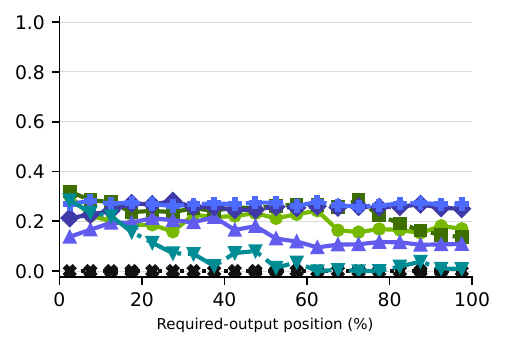}
    \caption{Heterogeneous table permutation}\label{fig:position-all-table-heterogeneous}
  \end{subfigure}\hfill
  \begin{subfigure}[t]{0.32\textwidth}
    \centering\includegraphics[width=\linewidth]{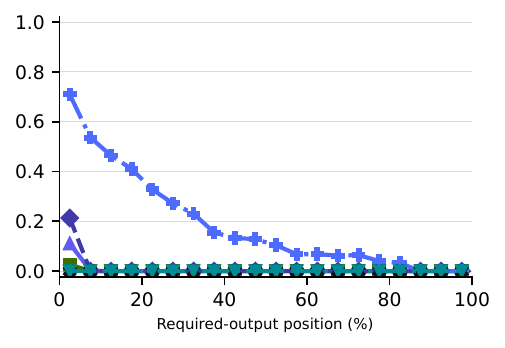}
    \caption{Table KV lookup}\label{fig:position-all-table-lookup}
  \end{subfigure}
  \caption{Accuracy across the required generation at 128K nominal tokens for all 12 task variants. Complete-grid curves average four difficulties and five document samples per setting after within-document binning; Olmo* has no 128K point. Figure columns compare input formats, while rows corresponds to tasks.}
  \label{fig:position-all}
\end{figure}

For numbered streams, the verifier uses the emitted numeric ID to match each output to its expected record. A missing ID therefore creates a localized zero without shifting later matches. For unnumbered streams, non-empty output lines are matched ordinally, so one omission can shift subsequent matches by design. Position plots (Figs.~\ref{fig:position},\ref{fig:position-all}) map expected record indices into 20 equal normalized bins, average correctness within each document and bin, and then macro-average documents. This procedure prevents easy settings with more short records from dominating the earlier bins in a curve.


\section{Additional results}

\begin{table}[h]
  \centering
  \small
  \caption{Exact document completion at 128K nominal tokens. A document needs no repair only when every required output item is correct.}
  \label{tab:exact-completion}
  \begin{tabular}{lrr}
    \toprule
    Model & Complete documents & Rate \\
    \midrule
    Nemotron 3 Nano & \NemotronNanoExactCountSixtyFiveK{}/240 & \NemotronNanoExactSixtyFiveK{}\% \\
    Nemotron 3 Super & \NemotronSuperExactCountSixtyFiveK{}/240 & \NemotronSuperExactSixtyFiveK{}\% \\
    Qwen3.5 35B-A3B & \QwenThirtyFiveExactCountSixtyFiveK{}/240 & \QwenThirtyFiveExactSixtyFiveK{}\% \\
    Qwen3.5 122B-A10B & \QwenOneTwentyTwoExactCountSixtyFiveK{}/240 & \QwenOneTwentyTwoExactSixtyFiveK{}\% \\
    DeepSeek V4 Flash & \DeepSeekFlashExactCountSixtyFiveK{}/240 & \DeepSeekFlashExactSixtyFiveK{}\% \\
    Kimi Linear 48B-A3B & \KimiLinearExactCountSixtyFiveK{}/240 & \KimiLinearExactSixtyFiveK{}\% \\
    Falcon-H1 34B & \FalconHOneExactCountSixtyFiveK{}/240 & \FalconHOneExactSixtyFiveK{}\% \\
    \bottomrule
  \end{tabular}
\end{table}

\begin{figure}[p]
  \centering
  \captionsetup[subfigure]{font=scriptsize,skip=1pt}
  \includegraphics[width=0.82\textwidth]{figures/legend_models.pdf}\\[-3pt]
  \begin{subfigure}[t]{0.32\textwidth}
    \centering\includegraphics[width=\linewidth]{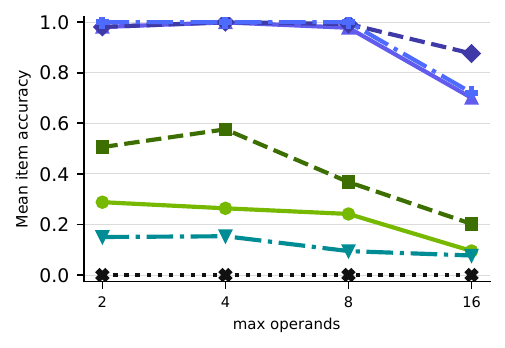}
    \caption{Arithmetic: ordered + IDs}\label{fig:difficulty-arithmetic-ordered}
  \end{subfigure}\hfill
  \begin{subfigure}[t]{0.32\textwidth}
    \centering\includegraphics[width=\linewidth]{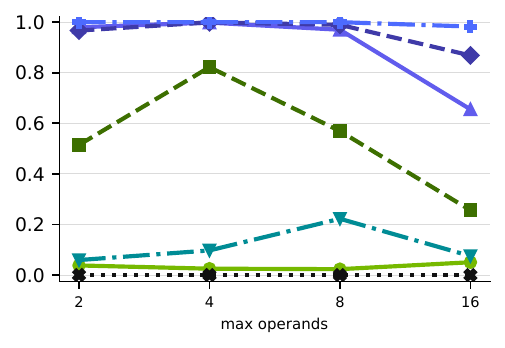}
    \caption{Arithmetic: shuffled + IDs}\label{fig:difficulty-arithmetic-shuffled}
  \end{subfigure}\hfill
  \begin{subfigure}[t]{0.32\textwidth}
    \centering\includegraphics[width=\linewidth]{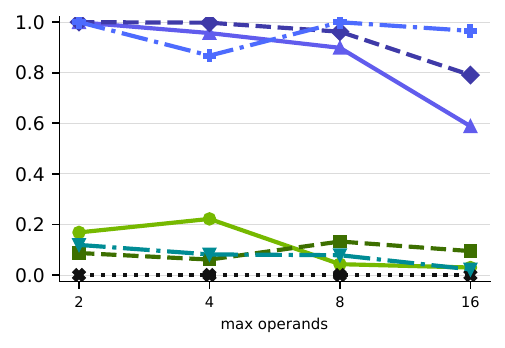}
    \caption{Arithmetic: no IDs}\label{fig:difficulty-arithmetic-noids}
  \end{subfigure}

  \begin{subfigure}[t]{0.32\textwidth}
    \centering\includegraphics[width=\linewidth]{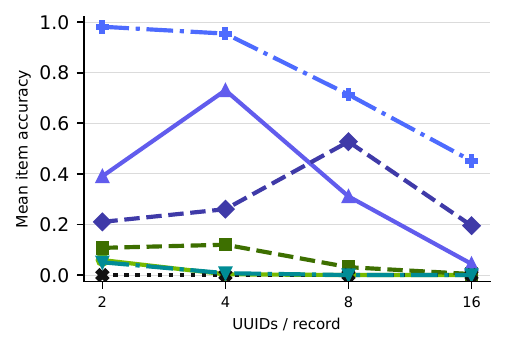}
    \caption{UUID sorting: ordered + IDs}\label{fig:difficulty-uuid-ordered}
  \end{subfigure}\hfill
  \begin{subfigure}[t]{0.32\textwidth}
    \centering\includegraphics[width=\linewidth]{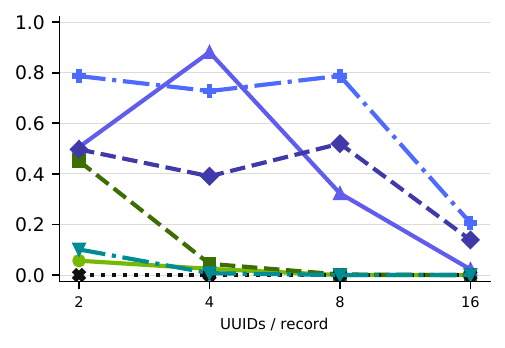}
    \caption{UUID sorting: shuffled + IDs}\label{fig:difficulty-uuid-shuffled}
  \end{subfigure}\hfill
  \begin{subfigure}[t]{0.32\textwidth}
    \centering\includegraphics[width=\linewidth]{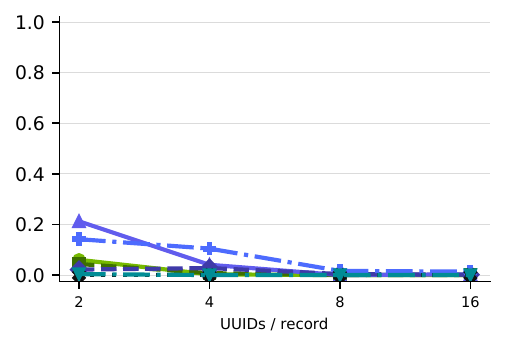}
    \caption{UUID sorting: no IDs}\label{fig:difficulty-uuid-noids}
  \end{subfigure}

  \begin{subfigure}[t]{0.32\textwidth}
    \centering\includegraphics[width=\linewidth]{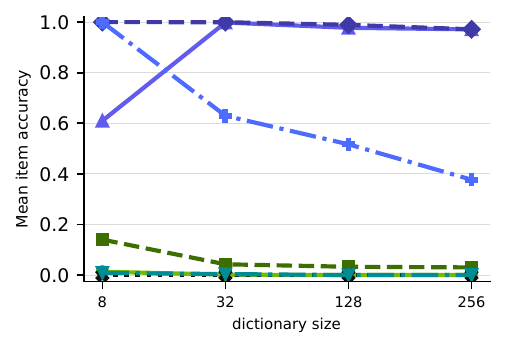}
    \caption{Lookup: ordered + IDs}\label{fig:difficulty-lookup-ordered}
  \end{subfigure}\hfill
  \begin{subfigure}[t]{0.32\textwidth}
    \centering\includegraphics[width=\linewidth]{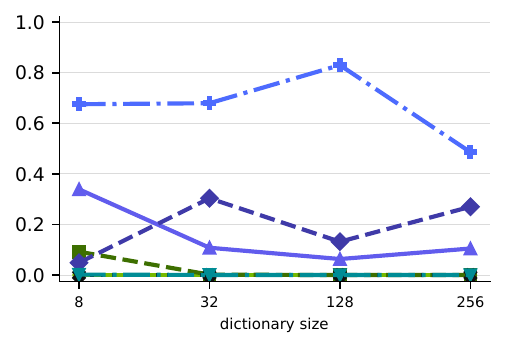}
    \caption{Lookup: shuffled + IDs}\label{fig:difficulty-lookup-shuffled}
  \end{subfigure}\hfill
  \begin{subfigure}[t]{0.32\textwidth}
    \centering\includegraphics[width=\linewidth]{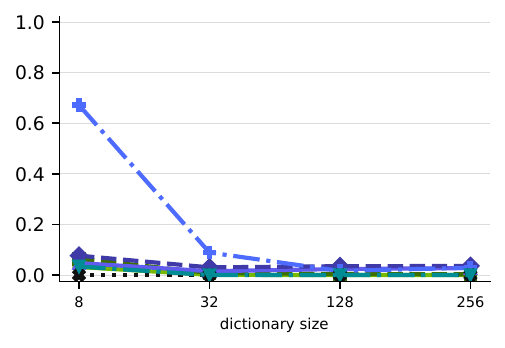}
    \caption{Lookup: no IDs}\label{fig:difficulty-lookup-noids}
  \end{subfigure}

  \begin{subfigure}[t]{0.32\textwidth}
    \centering\includegraphics[width=\linewidth]{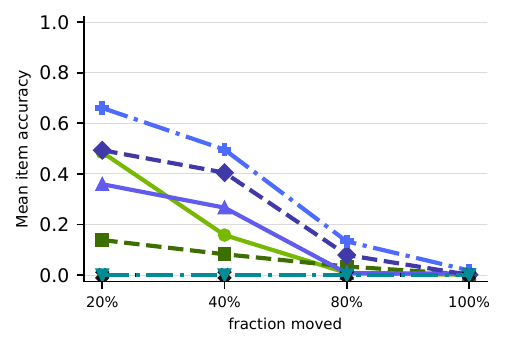}
    \caption{Homogeneous table permutation}\label{fig:difficulty-table-homogeneous}
  \end{subfigure}\hfill
  \begin{subfigure}[t]{0.32\textwidth}
    \centering\includegraphics[width=\linewidth]{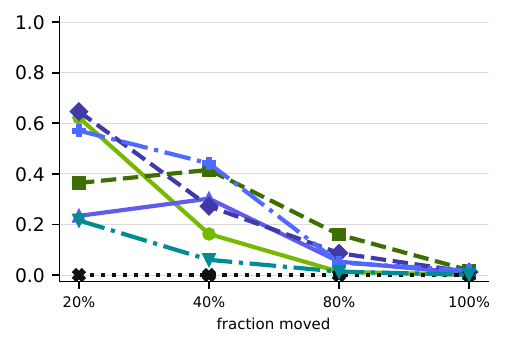}
    \caption{Heterogeneous table permutation}\label{fig:difficulty-table-heterogeneous}
  \end{subfigure}\hfill
  \begin{subfigure}[t]{0.32\textwidth}
    \centering\includegraphics[width=\linewidth]{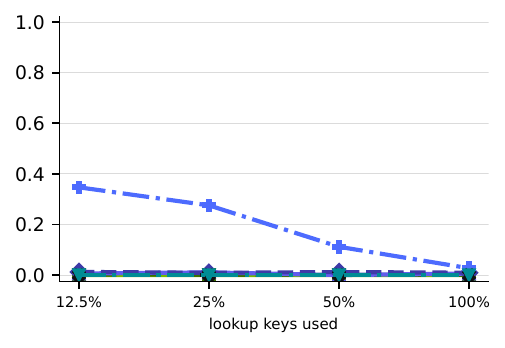}
    \caption{Table KV lookup}\label{fig:difficulty-table-lookup}
  \end{subfigure}
  \caption{All local-difficulty curves at 128K nominal tokens. Olmo* has no 128K point. These figures complement what is shown in Fig.~\ref{fig:overall-difficulty} which averages over all context lengths and formats.}
  \label{fig:difficulty}
\end{figure}

\begin{figure}[p]
  \centering
  \includegraphics[width=0.92\textwidth]{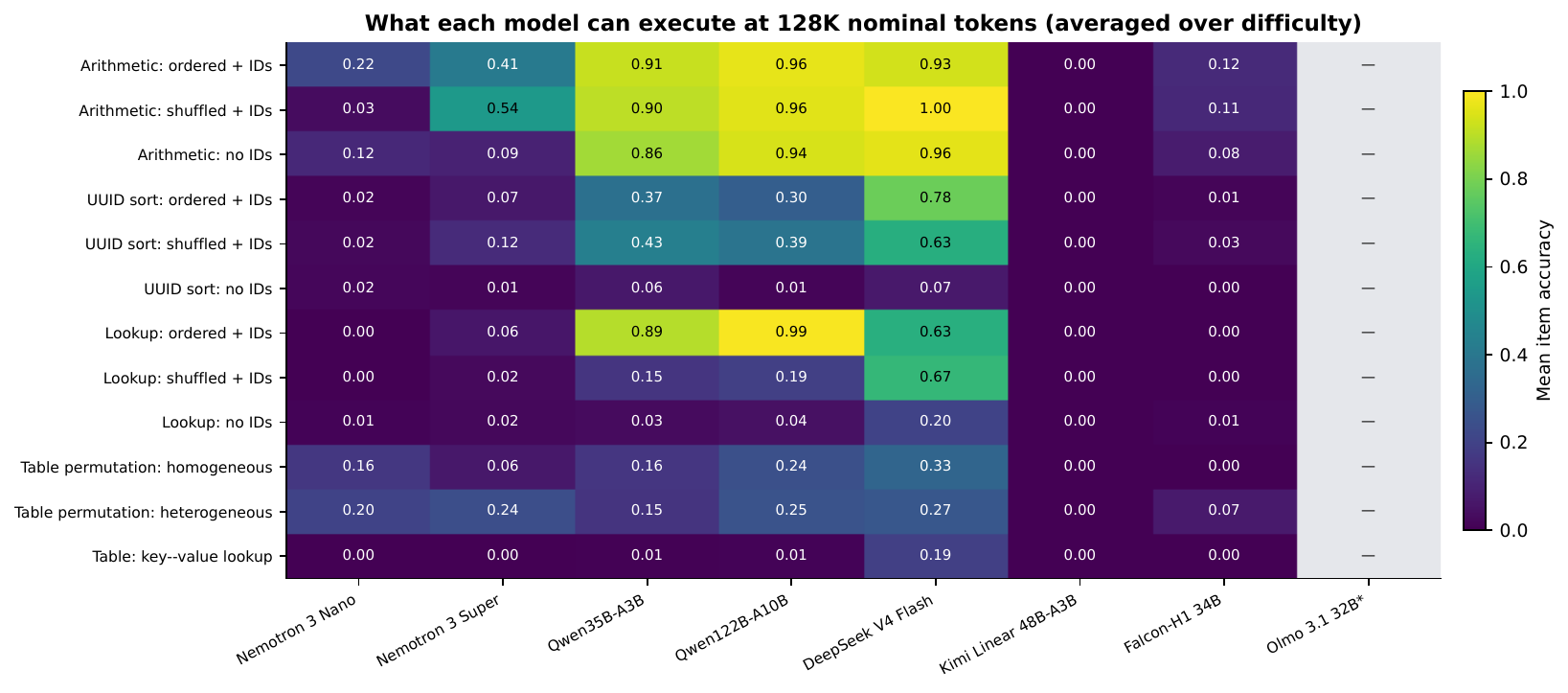}
  \caption{Available task-model results at 128K nominal tokens. Complete-grid cells average four difficulties and five document samples per setting; gray dashes mark missing Olmo* cells. This view complements Fig.~\ref{fig:difficulty}: the heatmap compares absolute performance while the curves show difficulty sensitivity.}
  \label{fig:task-heatmap}
\end{figure}

Figure~\ref{fig:task-heatmap} reports every task/format at the longest context length (128K).

\section{Reproducibility and Release}
\label{sec:reproducibility}
The frozen benchmark grid contains 4 tasks, 3 input formats, 4 difficulty settings, 6 nominal input-plus-output horizons, and 5 document samples per task: $4\times 3\times4\times6\times5=1{,}440$ documents. Each of seven primary models covers this identical grid. Olmo covers four full tiers through 32K; its context limit prevents evaluation at longer horizons. At the largest tier, the fixed \texttt{tiktoken} \texttt{cl100k\_base} preparation tokenizer estimates \MeanInputTokensSixtyFiveK{} mean prompt tokens and \MeanRequiredOutputTokensSixtyFiveK{} mean required-output tokens. Observed model-specific input counts differ because tokenizers differ. 

Inference uses greedy sampling with zero temperature. Every prompt explicitly instructs the model not to think. We set explicit thinking/reasoning to off or none when supported and otherwise use the lowest available effort (low). All plotted response-usage records report zero reasoning tokens. Outputs are permitted up to 1.5 times the target input budget.

Upon acceptance, we will release a fully seeded data generator, exact verifier, and evaluation scripts.

\section{Limitations}
LongTransduction is intentionally synthetic and deterministic. It does not test planning, tool selection, multi-turn interaction, recovery, permissions, human escalation, or direct safety behavior. Its records resemble enterprise data-processing primitives but are not a realistic enterprise environment. Positional scoring of unnumbered tasks also makes omissions cascade by design; this measures format fragility as well as local competence. However, this synthetic setting mimics fundamental primitives seen in real-world environments.

\begin{figure}[t]
  \centering
  \includegraphics[width=0.66\textwidth]{figures/legend_models.pdf}\\[-2pt]
  \begin{subfigure}[t]{0.49\textwidth}
    \centering
    \includegraphics[width=\linewidth]{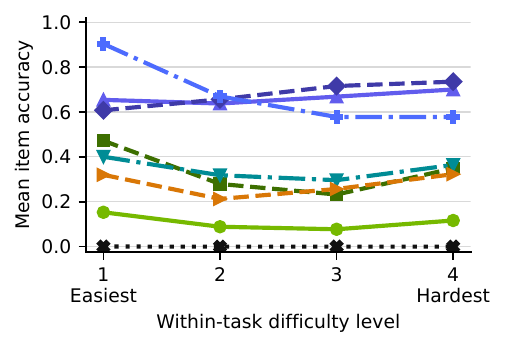}
    \caption{Arithmetic}
    \label{fig:variable-lookup-difficulty}
  \end{subfigure}
  \begin{subfigure}[t]{0.49\textwidth}
    \centering
    \includegraphics[width=\linewidth]{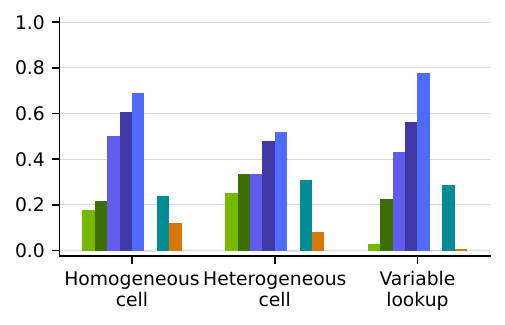}
    \caption{Table Transform}
    \label{fig:protocol-table}
  \end{subfigure}
  \caption{Fig.~\ref{fig:variable-lookup-difficulty} corresponds to the figures shown in Fig.~\ref{fig:difficulty}. Fig.\ref{fig:protocol-table} corresponds to the figures shown in Fig.~\ref{fig:protocol}.}
  \label{fig:overall-difficulty-and-protocol}
\end{figure}

\end{document}